# Optimal Prediction Intervals for Macroeconomic Time Series Using Chaos and NSGA–II

Vangala Sarveswararao[2], Vadlamani Ravi[1*][0000-0003-0082-6227], Sheik Tanveer Ul Huq[2]

[1]Centre of Excellence in Analytics,
Institute for Development and Research in Banking Technology,
Castle Hills Road No. 1, Masab Tank, Hyderabad-500057, India
[2]SCIS, University of Hyderabad, Hyderabad-500046, India

**Abstract**

In a first-of-its-kind study, this paper proposes the formulation of constructing prediction intervals (PIs) in a time series as a bi-objective optimization problem and solves it with the help of Nondominated Sorting Genetic Algorithm (NSGA-II). We also proposed modeling the chaos present in the time series as a preprocessor in order to model the deterministic uncertainty present in the time series. Even though the proposed models are general in purpose, they are used here for quantifying the uncertainty in macroeconomic time series forecasting. Ideal PIs should be as narrow as possible while capturing most of the data points. Based on these two objectives, we formulated a bi-objective optimization problem to generate PIs in 2-stages, wherein reconstructing the phase space using Chaos theory (stage-1) is followed by generating optimal point prediction using NSGA-II and these point predictions are in turn used to obtain PIs (stage-2). We also proposed a 3-stage hybrid, wherein the 3$^{rd}$ stage invokes NSGA-II too in order to solve the problem of constructing PIs from the point prediction obtained in 2$^{nd}$ stage. The proposed models when applied to the macroeconomic time series, yielded better results in terms of both prediction interval coverage probability (PICP) and prediction interval average width (PIAW) compared to the state-of-the-art Lower Upper Bound Estimation Method (LUBE) with Gradient Descent (GD). The 3-stage model yielded better PICP compared to the 2-stage model but showed similar performance in PIAW with added computation cost of running NSGA-II second time.

## 1. Introduction

Even though neural networks (NNs) have shown impressive performance in terms of prediction accuracy, in many real-world applications, however, the uncertainty of each prediction must also be quantified as there is a large downside to making an incorrect prediction in areas such as finance, weather, traffic, manufacturing, energy networks and prognostics. Krzywinski & Altman [1] and Gal [2] experimented on NNs to meet this requirement. Prediction intervals (PIs) will communicate with us directly by offering a lower and an upper bound for each prediction and this information helps us make better-informed decisions.

---

* Corresponding Author; Tel.: +914023294310; Fax: +914023534551; E-mail: rav_padma@yahoo.com

A variety of approaches have been developed, ranging from fully Bayesian NNs [3] to interpreting dropout as performing variational inference [4]. However, these methods require either strong assumptions or high computational demands for running them. Khosravi et al. [5] developed the Lower Upper Bound Estimation (LUBE) method, which uses NN to generate PIs and its loss function is incompatible with Gradient Descent (GD) for training. Later, Pearce et al. [6] developed a new loss function which is compatible with GD outperformed the LUBE based state-of-the-art methods.

In this work, we formulate the construction of PIs as a multi-objective optimization problem and solve it by NSGA-II [7]. PIs should capture as many data points as possible, while narrowing the width of the interval. Accordingly, quality of PIs is mainly assessed by using the measures viz., prediction interval coverage probability (PICP) and prediction interval average width (PIAW) [8] [9] [10]. We utilized these two objectives and invoked NSGA-II to quantify the uncertainty for each prediction. First, we estimated the point predictions using NSGA-II by minimizing forecasting errors as one objective and directional symmetry as another objective and then we formulated two equations to compute both lower and upper bounds using these point predictions. These equations involve two random numbers. In the 2-stage method, we used the grid search to get two optimal random numbers to be used in the equation. However, the 3-stage method is proposed with two variants, where we invoked NSGA-II with PICP as one objective and PIAW as another to obtain the optimal combination of random numbers. The chief advantages of the proposed methods are its intuitive objectives and low computational cost compared to LUBE method as it requires training an NN with the help of evolutionary algorithms.

The rest of the paper is organized as follows: Section 2 presents the literature review; Section 3 presents the overview the techniques used; the Section 4 presents our proposed model in detail; Section 5 presents experimental methodology; Section 6 presents a discussion of the results and finally section 6 concludes the paper and presents future directions.

## 2. Literature survey

Tibshirani [11], Papadopoulos et al. [8] and Khosravi et al. [9] are the pioneers who worked on quantifying uncertainty in regression with neural networks. While the first one focuses on Confidence Intervals (CIs), the latter works specifically focus on forecasting prediction intervals. They have presented three primary methods:

- **Delta Method** [12] which follows the theory for constructing CIs used by non-linear regression models and this method is computationally expensive as it requires the use of Hessian Matrix.
- **Mean-Variance Estimation** (MVE) (Nix and Weigend [13]) uses a neural network with two output nodes, one representing mean and the other representing the variance of the normal distribution, which allows the estimation of the variance of data noise. In this method, they have used Negative Log-likelihood of the prediction distribution of the given data as a loss function.
- **Bootstrap method** (Heskes [14]) trains multiple neural networks on different resampled versions of the training data with different initializations of parameters. It can be easily combined with MVE to estimate total variance. Lakshninarayan et al.[15] improved the work of Heskes [14] by ensembling individual MVE Neural Networks by resampling the training set and including adversarial training examples, which they refer as MVE Ensemble.

Lower Upper Bound Estimation (LUBE) method was developed by Khosravi et al. [5] based on the principle of High-Quality Prediction Intervals. LUBE method used a neural network to construct prediction intervals and the parameters of the NN were estimated using the Simulated Annealing algorithm. Various non-gradient methods such as Genetic Algorithms (Ak et al. [16]), Gravitational Search Algorithms (Lian et al. [17]), Particle Swarm Optimization (Galvan et al. [10]; Wang et al. [18]), Extreme Learning Machines (Sun et al. [19]), and Artificial Bee Colony Algorithms (Shen et al. [20]) were proposed to improve the predictions. LUBE method has been used in many applications including predicting energy load (Wan et al. [21]; Quan et al. [22]), wind speed ([18] [16]), landslide displacement ([17]), solar energy ([10]) and others.

Works in PIs for financial time series include Muller and Watson's [23] Robust Bayes PI algorithm to forecast long term prediction intervals for economic growth over a horizon of 10- 75 years and Chudy et al. [24] works on computational adjustments to the Bayesian & bootstrapping methods to improve coverage probability for eight macroeconomic indicators. Sarveswararao and Ravi [25] used LSTM architecture instead of NN in the LUBE method as LSTM are known for their capability to learn complex sequential information and outperformed the LUBE method with GD for constructing PIs for CPI inflation. Several works were published on inflation forecasting as it plays an important role in monetary policy formulation in many countries [26] [27] [28] [29] and these works computed point prediction to inflation without any PIs. Pradeepkumar & Ravi [31] worked on the FOREX rate predictions and improved the accuracy by modelling chaos before applying any forecasting algorithm and Ravi et al. [30] proposed hybrids of Neural Networks + Evolutionary algorithms; Quantile regression random forest [32], and Multivariate Regression Splines [33] for forecasting the same. Some works on generating PIs without NN include Kumar et al. [34]'s MapReduce based Fuzzy very fast decision tree and Ravi et al. [35]'s hybrid consisting of support vector machine & quantile regression random forest. Krishna and Ravi [36] presented a comprehensive survey of Evolutionary computing (EC) algorithms applied to Customer Relationship Management (CRM), which focused on using various EC algorithms to solve simple to complex analytical CRM tasks, which turn out to be data mining tasks. So far, Evolutionary Multi-Objective (EMO) algorithms were conspicuous by their absence in generating PIs. Instead, they were employed to train NNs which in turn generated PIs in case of LUBE method. Our work mainly focusses on constructing PIs for an important macroeconomic time series, namely, Consumer Price Index using NSGA-II without any NNs.

## 3. Overview of the techniques used

### 3.1. Chaos theory

In the late 1800s, the theory of chaos was proposed by Poincare and later extended by Lorenz [37] in 1963 to deal with unpredictable complex nonlinear systems [38]. A chaotic system is deterministic, dynamic and evolves from the given initial conditions and it can be described by trajectories in the state space. As the governing equations are not known ahead for a chaotic system, the state space is represented by phase space, which can be reconstructed from the original time series. This reconstructed phase space provides a multi-dimensional view of the original time series [38]. Packard et al. [39] proposed a method to reconstruct the phase space using the method delays, using which for a time series $X_i$ where i = 1, 2, 3, .., n, the phase space can be constructed by a m-dimensional vector as in Eq. (1):

$$Y_i = (x_i, x_{i+\tau}, x_{i+2\tau}, \ldots, x_{i+(m-1)\tau}) \tag{1}$$

Where $\tau$ is the delay time or the lag of the system and $m$ is the embedding dimension to reconstruct the phase space. After the reconstruction of phase space, the time series problem gets converted into multi-input single-output (MISO) prediction problem, which can be modelled by methods ranging from linear models to deep neural networks.

### 3.2. Rosenstein's method

Rosenstein's algorithm [40] (see Fig. 1) estimates the largest Lyapunov Exponent ($\lambda$) [41] from the given time series as shown as in the figure 1. If $\lambda \geq 0$ then chaos is present; otherwise chaos is absent in a given time series.

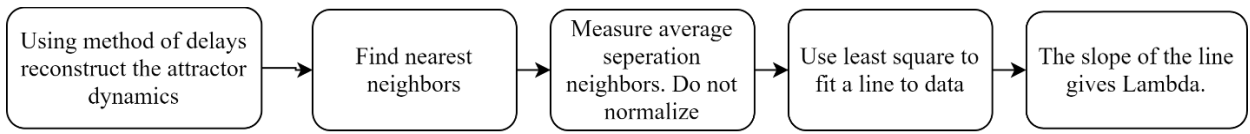

**Fig. 1**. Rosenstein's method to calculate $\lambda$

## 3.3. Cao's method

Cao proposed a method [42] to find out the minimum embedding dimension for a given times series. Let $X = (x_1, x_2, x_3, x_4, \ldots, x_N)$ be a time series. In the phase space, the time series can be reconstructed as in Eq. (2):

$$Y_i = (x_i, x_{i+\tau}, x_{i+2\tau}, \ldots, x_{i+(m-1)\tau}) \qquad (2)$$

Where, $Y_i$ is the i[th] reconstructed vector, and $\tau$ is the time delay.

## 3.4 Multi-objective evolutionary algorithms

MOEAs were proposed to solve optimization problems involving two objectives. The MOEAs preserve the solutions which are non-dominating, while progressing algorithmically towards the optimal Pareto front and maintaining diversity in the optimal front. The decision maker can select the solutions from this optimal Pareto front [43] to solve his problem. Deb [44], Mukhopadhyay et al. [36, 37] and Coello Coello et al. [43] proposed some of the works on MOEAs.

### 3.4.1 Non-dominated Sorting Genetic Algorithm – II (NSGA - II)

NSGA [109] is one of the famous and effective algorithms for solving multi-objective optimization problems. Still, it attracted criticism for its large computational complexity, absence of elitism, requirement of niching and for choosing optimal $\sigma_{share}$. Deb [7] modified the original NSGA method to use elitism, a better sorting algorithm & the number of parameters to be chosen ahead and he named this modified version as NSGA-II.

For NSGA-II, the population is initialized just like NSGA followed by sorting it based on the objectives into various fronts in a hierarchical way. The initial front has contains non-dominated set of individuals in the whole populations and the second front is only dominated by first, the third front is dominated only by first and second, and the same goes for all. After the sorting is over, crowding distance is calculated for each candidate in the front. It is a measure of closeness of each candidate in the objective space to other members in the population.

The fitness value or rank of 1 is given to members in the first front, and two is given to members in the second front and so on. Based on their rank and crowding distance, non-dominated members of the population are selected for further processing. Offspring population (Qt) is generated from the parent population (Pt) by crossover & mutation operations, and they are combined with parent populations to select best N members based on non-dominated sorting. This process of generating a new population goes on until we reach a terminated condition. At every generation, all the best solutions from the previous generations are passed on, which takes care of elitism. A new generation (Pt+1) obtained by merging each front (Fi) until the number of individuals in the selection exceeds the population, in that case, we consider the crowding distance to select best ones from that front until we reach N (see Fig. 2).

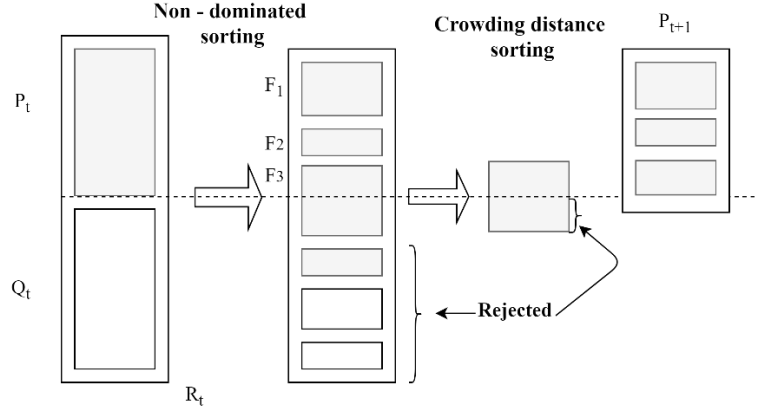

**Fig. 2.** Non-dominated Sorting Genetic Algorithm (NSGA-II)

## 4. Proposed models

### *4.1. Formulating multi-objective optimization problem*

We can formulate generating PIs as a bi-objective optimization problem as follows:

*Minimize $O_1$ = Symmetric Mean Absolute Percentage Error (SMAPE) and*

*Maximize $O_2$ = Directional Symmetry Statistic (DS)* **[47]**

Where SMAPE and DS are defined in Eqs. (3) and (4) respectively.

$$SMAPE = \frac{100\%}{n} \sum_{t=1}^{n} \frac{|Y_t - Y\hat{}_t|}{(|Y_t| + |Y\hat{}_t|)/2} \qquad (3)$$

$$DS(A,F) = \frac{100}{n-1} \sum_{i=2}^{n} d_i$$
$$d_i = \begin{cases} 1, & if\ (Y_i - Y_{i-1})(Y\hat{}_i - Y\hat{}_{i-1}) > 0 \\ 0, & otherwise \end{cases} \qquad (4)$$

Where $Y_t$ is the actual value and $Y_t\hat{}$ is the predicted value and n is the number of data points.

In forecasting financial time series, minimizing the SMAPE is important as it ignores the scale of underlying data while minimizing the average prediction errors. And the same important has to be given to Directional symmetry statistic as it helps in predicting the movement of the series.

## 4.2. Proposed NSGA-II based 2-stage model

Let $Y = (y_1, y_2, y_3, y_4, \ldots, y_N)$ be a time series of size N. Prediction intervals are constructed using the 2-stage model as follows:

### Stage 1: Reconstructing the phase space

- Check whether Y contains chaos. If the test is positive then rebuild the phase space of Y with the help of time delay $(\tau)$ and embedding dimension $(m)$.
- Partition $Y$ into $Y_{Train} = \{y_t; t = \tau m + 1, \tau m + 2, \ldots, h\}$ and $Y_{Test} = \{y_t; t = h + 1, h + 2, \ldots, N\}$.

### Stage 2: Constructing Optimal Prediction Intervals

Obtain optimal coefficients for auto-regression model using NSGA-II, where the auto-regression coefficients are the decision variables which are in the range (-0.5,0.5) with SMAPE and DS as the objective functions to be minimized and maximized respectively. Now, using those coefficients obtain the predictions for both training & test set. These predictions are then used for computing the lower and upper bounds (a.k.a. prediction intervals) for our series with the help of Eq. (5) and (6).

$$\hat{Y}_t = \alpha_0 + \alpha_1 y_{t-\tau} + \alpha_2 y_{t-2\tau} + \alpha_3 y_{t-3\tau} + \cdots + \alpha_m y_{t-\tau m}$$

Where $\hat{Y}_t$ is point prediction at time t and $(\alpha_0, \alpha_1, \alpha_2, \ldots, \alpha_m)$ are decision variables / auto-regression coefficients, while $(y_{t-\tau}, y_{t-2\tau}, y_{t-3\tau}, \ldots, y_{t-m\tau})$ are input features after chaos modelling.

$$\hat{Y}_{lower} = \hat{Y}_t - r_1 * \sigma(\hat{Y}_{train}) \quad (5)$$

$$\hat{Y}_{upper} = \hat{Y}_t + r_2 * \sigma(\hat{Y}_{train}) \quad (6)$$

Where $r_1 \text{ and } r_2$ are two random numbers following uniform distribution between (0,1) and we applied grid search to find their values in order to obtain optimal PICP and PIAW, $(\hat{Y}_{Lower}^t, \hat{Y}_{Upper}^t)$ are the lower and upper bounds for $\hat{Y}_t$, $and$ $\sigma(\hat{Y}_{train})$ is the standard deviation of the point predictions of the training set. The schematic for the 2-satge model is depicted in Fig 3.

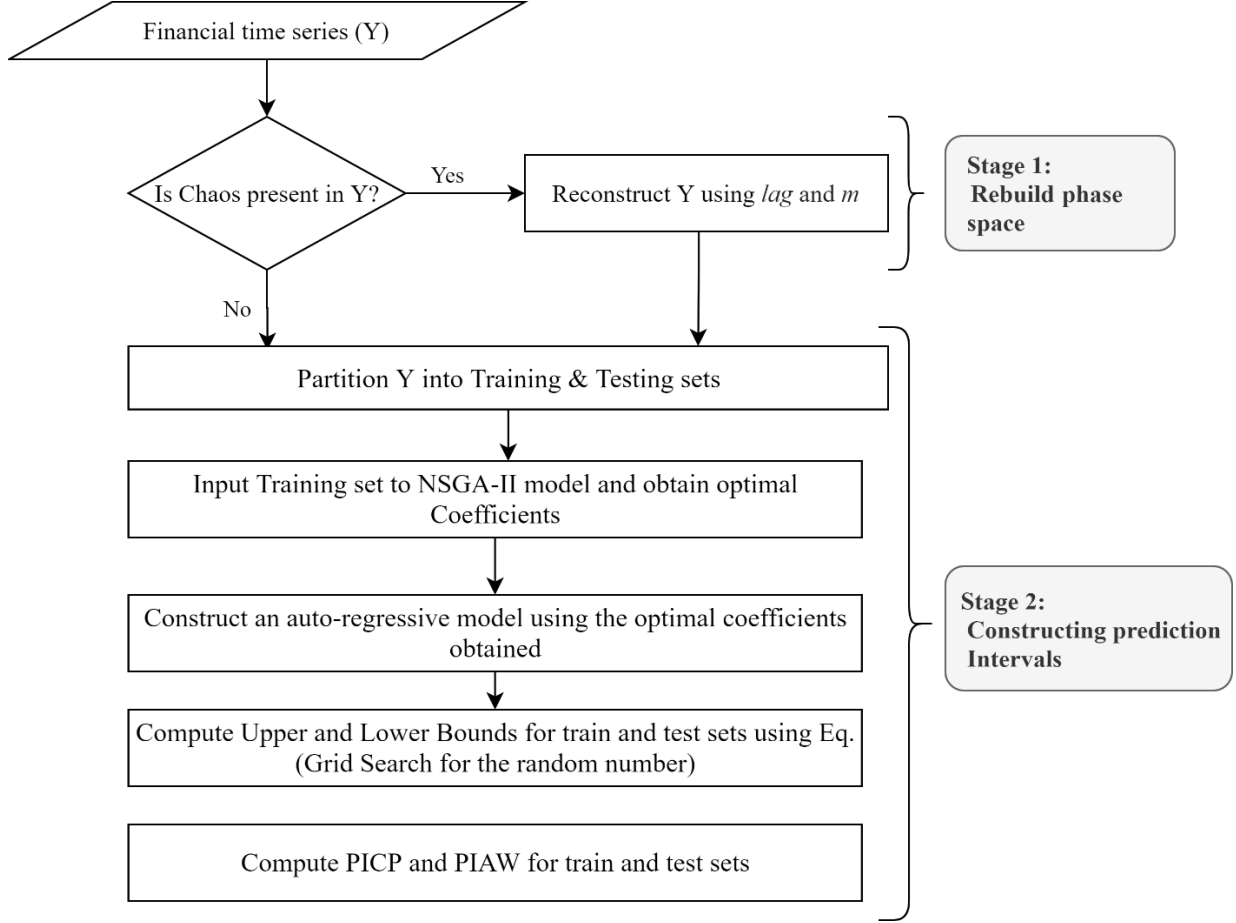

**Fig. 3.** Schematic of the 2-stage model

## *4.3. Proposed NSGA-II based 3-stage model*

Let $Y = (y_1, y_2, y_3, y_4, \ldots, y_N)$ be a time series of size N. Prediction intervals for CPI are constructed using the three-stage model as follows. Here NSGA-II is invoked twice.

*Stage 1: Reconstructing the phase space*

- Check whether Y contains chaos. If the test is positive then rebuild the phase space of Y with the help of time delay ($\tau$) and embedding dimension ($m$).
- Partition $Y$ into $Y_{Train} = \{y_t; t = \tau m + 1, \tau m + 2, \ldots, h\}$ and $Y_{Test} = \{y_t; t = h + 1, h + 2, \ldots, N\}$.

*Stage 2: Obtaining Optimal Auto regression coefficients*

Obtain optimal coefficients for auto-regression model using NSGA-II, where the auto-regression coefficients are the decision variables which are in the range (-0.5,0.5) with SMAPE and DS as the objective functions to be minimized and maximized respectively. Now, using those coefficients obtain the predictions for both training & test set. These predictions are then used for getting lower & upper bounds for our series (a.k.a. prediction intervals) with the help of Eq. (6) and (7).

$$\hat{Y}_t = \alpha_0 + \alpha_1 y_{t-\tau} + \alpha_2 y_{t-2\tau} + \alpha_3 y_{t-3\tau} + \cdots + \alpha_m y_{t-\tau m}$$

Where $\hat{Y}_t$ is point prediction at time t and $(\alpha_0, \alpha_1, \alpha_2, \ldots, \alpha_m)$ are decision variables / auto-regression coefficients. $(y_{t-\tau}, y_{t-2\tau}, y_{t-3\tau}, \ldots, y_{t-m\tau})$ are input features after chaos modelling.

### Stage 3: Constructing prediction intervals

In this final stage, instead of performing the grid search to find the optimal random numbers $r_1$ and $r_2$, we invoke NSGA-II which optimizes PICP and PIAW explicitly.

$$\hat{Y}_{lower} = \hat{Y}_t - r_1 * \sigma(\hat{Y}_{train}) \qquad (6)$$
$$\hat{Y}_{upper} = \hat{Y}_t + r_2 * \sigma(\hat{Y}_{train}) \qquad (7)$$

Where $r_1$ and $r_2$ are the decision variables in the range (0,1) and NSGA-II will find optimal values for decision variables. $(\hat{Y}^t_{lower}, \hat{Y}^t_{upper})$ are lower and upper bounds for $\hat{Y}_t$. $\sigma(\hat{Y}_{train})$ is the standard deviation of the point predictions of the training set.

Thus, we apply NSGA-II in both 2nd and 3rd stages. In the 3rd stage, we have two variants (i) single random number (ii) two distinct random numbers.

In the first variant, we applied NSGA-II to get a single random number so that the PI's are obtained with the assumption $r_1 = r_2 = r$. whereas the 2nd variant assumes that they are distinct. The optimization problem is as follows:

*Maximize $O_3$ = PICP and*

*Minimize $O_4$ = PIAW*

Where $PICP = \frac{c}{n}$, $PIAW = \frac{1}{n}\sum_{i=1}^{n} \hat{y}U_i - \hat{y}L_i$ and c is the number of data points captured out of total n points by prediction interval and $<\hat{y}L_i, \widehat{yU_i}>$ are predicted lower and upper bounds.

The schematic for the 3-stage model is depicted in Fig. 4.

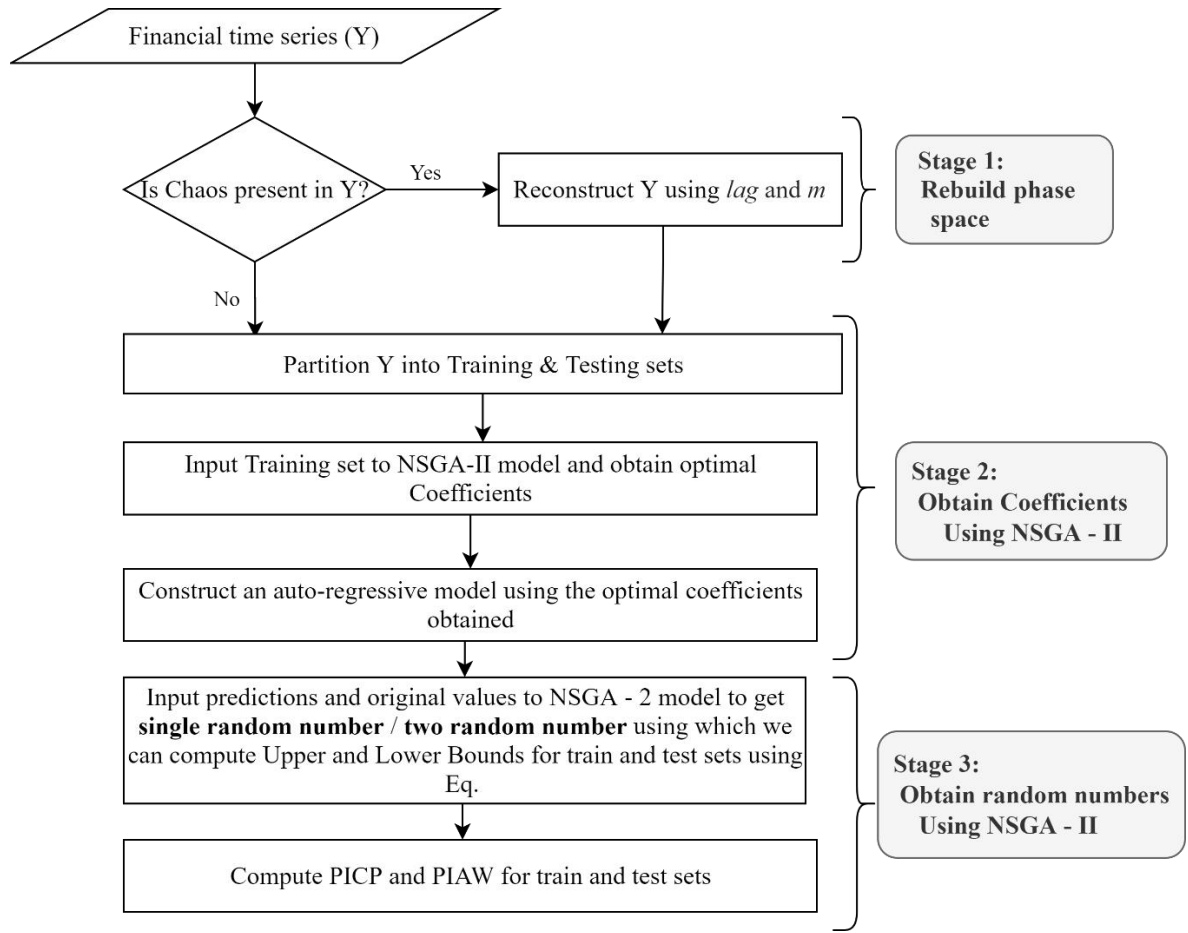

**Fig. 4.** Schematic of the 3-stage model

# 5. Experimental Design

## 5.1 Data sets used

We used the Consumer Price Index (CPI) Inflation data of Food & Beverages, Fuel & Light and Headline from the Ministry of Statistics and Programme Implementation (MoSPI), Government of India. The dataset contains monthly inflation starting from January 2012 to December 2018 presented in Fig. 5. Summary statistics of this dataset presented in Table 1 indicates that inflation in India is highly volatile. Table 2 presents the Lyapunov exponent values as well as the minimum embedding dimension for all CPI series.

**Table 1:** Summary Statistics of the Datasets.

| Indicator | CPI Food & Beverages | CPI Fuel & Light | CPI Headline |
|---|---|---|---|
| *Mean* | 6.23 | 6.19 | 6.23 |
| *Standard deviation* | 4.08 | 2.35 | 2.71 |
| *Maximum* | 14.45 | 13.13 | 11.16 |
| *Minimum* | -1.69 | 2.49 | 1.46 |

**Table 2:** Chaotic Modelling of Macroeconomic Variables

| Series | Lyapunov Exponent | Delay Time | Embedding Dimension |
|---|---|---|---|
| CPI Headline Inflation | 0.074 | 1 | 8 |
| CPI Food & Beverages Inflation | 0.062 | 1 | 8 |
| CPI Fuel & Light Inflation | 0.060 | 1 | 7 |

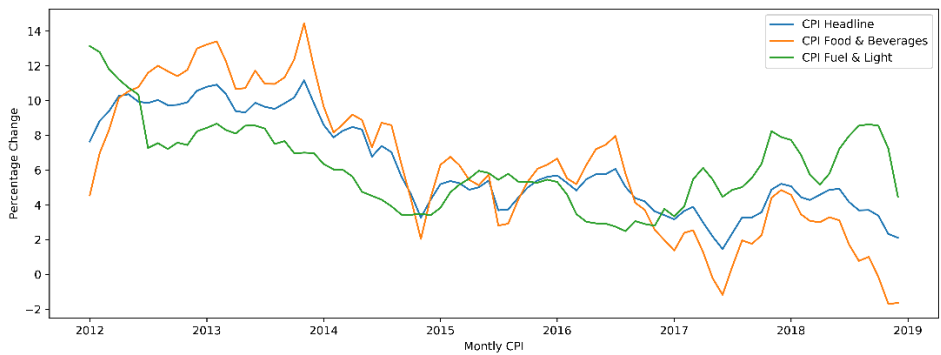

**Fig. 5.** CPI Headline & its Components Monthly inflation from 2012M01 to 2018M12.

We considered last 6 months' data to be the test and previous data to be the training set. (see Table 3)

**Table 3:** Train – Test Split for Data sets

| Series | Train set | Test set |
|---|---|---|
| CPI Headline | Jan'12 – Jun'18 | Jul'18 – Dec'18 |
| CPI Food & Beverages | Jan'12 – Jun'18 | July'18 – Dec'18 |
| CPI Fuel & Light | Jan'12 – Jun'18 | July'18 – Dec'18 |

*5.2 Tools and techniques used*

The proposed model execution is carried out using *Ubuntu 18.04 LTS* Platform with 32 GB RAM and 1 TB HDD. However, it can also be carried out on other platforms. Table 4 catalogues various tools employed in this study.

**Table 4:** Tools and Techniques used for Proposed Model

| Technique used | Used for | Tool used |
|---|---|---|
| Rosenstein's Method | Calculating Lyapunov exponent | Python |
| Cao's Method | Finding minimum embedding dimension | R |
| Auto Correlation Function | Finding Optimal lag/time delay | Python |
| NSGA-II | Obtaining Coefficients of AR model | Python |

*5.3 Performance measures used*

We have considered two metrics used in PIs literature for measuring the performance of the prediction intervals, namely, PICP and PIAW.

**Prediction Interval Coverage Probability** (PICP) gives us the percentage of data points in between lower bound and upper bound a.k.a Prediction Interval as in Eq. (6):

$$PICP = \frac{c}{n} \quad (6)$$

Where c is the total number of points captured by the intervals and n is the total number of points.

**Prediction Interval Average Width** (PIAW) gives us the average width between the lower and upper bounds as in Eq. (7):

$$PIAW = \frac{1}{n}\sum_{i=1}^{n} \hat{y}U_i - \hat{y}L_i. \quad (7)$$

Where $\hat{y}U_i$ and $\hat{y}L_i$ are predicted upper & lower bounds.

All three inflation series turns out to be non-stationary when tested using the Augmented Dickey-Fuller Test [48]. We are not accounting for the seasonality in the series as CPI inflation is assumed to contain no seasonality [26]. CPI inflation series contains chaos as confirmed by the Lyapunov Exponent, so we have to model chaos before embarking any forecasting algorithm.

# 6. Results and Discussion

The best performing hyperparameters for all the experiments in this study are presented in Table 8. We have compared our proposed model results with LUBE method with GD as their model was outperforming all non-gradient based training algorithms for LUBE method to construct prediction intervals. All the three proposed models outperformed the LUBE + GD and LUBE + LSTM in terms of both PICP and PIAW for constructing PI for CPI inflation data. We ran the experiments 20 times with 20 different seeds to remove the effect of the seed value. Accordingly, we presented the mean and standard deviation of PICP and PIAW corresponding to the best solution of the population over 20 experiments in Tables 5, 6 and 7 in the form of $mean \pm std.dev$.

## 6. 1. CPI Food and beverages inflation

Table 5 presents the PICP and PIAW values using two-stage model (Chaos + NSGA-II{SMAPE, DS}), and three-stage model (Chaos + NSGA-II {SMAPE, DS} + NSGA-II {PICP, PIAW}). According to the results, the performance of 2-stage model is improved 2x compared to LUBE + GD & LUBE + LSTM in terms of PICP metric and showed similar performance in terms of PIAW metric. By running NSGA-II second time with PICP and PIAW as objectives in the 3-stage model, the performance improved with respect to PICP metric but not in PIAW metric and it comes with an extra computational cost of running NSGA-II compared to single run using 2-stage model. Fig. 6 depicts that the predicted values of the two-stage model with LUBE + GD model, and it reveals that NSGA-II based 2-stage model could yield very closer prediction intervals compared to LUBE + GD method for CPI Food & Beverages. Fig. 7 compares the performance of the two variants of the 3-stage model.

**Table 5:** Results for CPI Food & Beverages inflation.

| Method | PICP | PIAW |
| --- | --- | --- |
| LUBE Method with GD | 0.44 ± 0.17 | 2.60 ± 1.16 |
| LUBE Method with LSTM [25] | 0.42 ± 0.11 | 2.37 ± 0.30 |
| NSGA-II based 2-stage model (Grid search) | 0.91 ± 0.11 | 2.59 ± 0.27 |
| NSGA-II based 3-stage model (Using single random number) | 1.00 ± 0.00 | 2.57 ± 0.44 |
| NSGA-II based 3-stage model (Using two random numbers) | 1.00 ± 0.00 | 2.86 ± 0.53 |

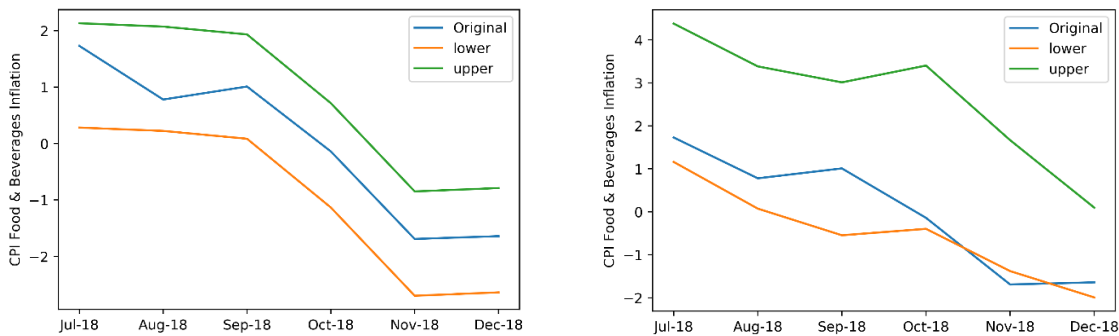

**Fig. 6.** PIs for CPI Food & Beverages using the 2-stage model (*left*) and LUBE + GD (*right*)

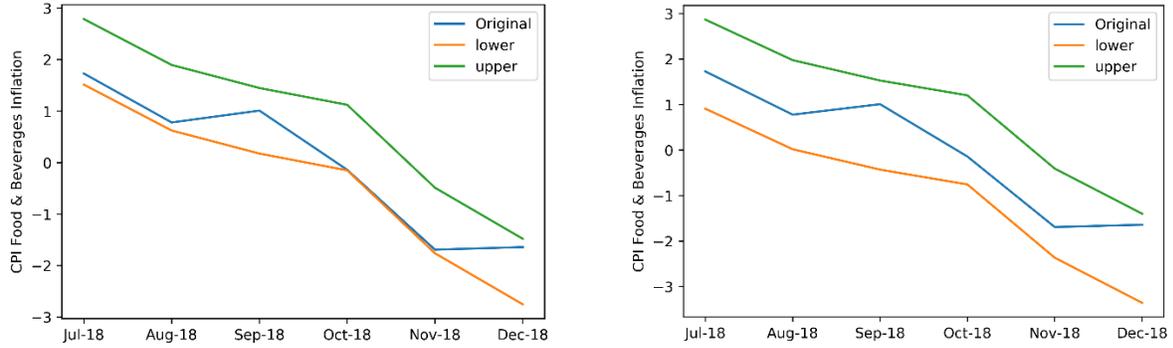

**Fig. 7.** PIs for CPI Food & Beverages using 3-stage model (single random no.) (*left*) and 3-stage model (two random no.) (*right*)

## 6.2. CPI Headline inflation

Table 6 presents the PICP and PIAW values using two-stage model (Chaos + NSGA-II{SMAPE, DS}), and three-stage model (Chaos + NSGA-II{SMAPE, DS} + NSGA-II{PICP, PIAW}). According to the results, the performance of NSGA-II based 2-stage model is improved by 9.4% compared to LUBE + GD in terms of PICP metric & 37% in terms of PIAW metric and when compared with LUBE + LSTM, PICP showed similar performance & PIAW got improved by 34%. The performance of NSGA-II based 3-stage model improved by 7.5% w.r.t PICP metric and 11.12% in terms of PIAW but the added benefits in performance come with an extra computational cost of running NSGA-II second time compared to single run using 2-stage model. Fig. 8 depicts the predicted values of the NSGA-II based 2-stage model with LUBE + GD model, and it reveals that 2-stage model could yield very closer prediction intervals compared to LUBE + GD method for CPI Headline inflation. Fig. 9 compares the performance of the two variants of the 3-stage model.

**Table 6:** Results for CPI Headline inflation

| Method | PICP | PIAW |
|---|---|---|
| LUBE Method with Gradient Descent | 0.85 ± 0.14 | 3.00 ± 1.18 |
| LUBE Method with LSTM [25] | 0.93 ± 0.13 | 2.54 ± 0.25 |
| NSGA-II based 2-stage model (Grid search) | 0.93 ± 0.20 | 1.89 ± 0.12 |
| NSGA-II based 3-stage model (Using single random number) | 1.00 ± 0.00 | 1.68 ± 0.53 |
| NSGA-II based 3-stage model (Using two random numbers) | 1.00 ± 0.00 | 1.74 ± 0.28 |

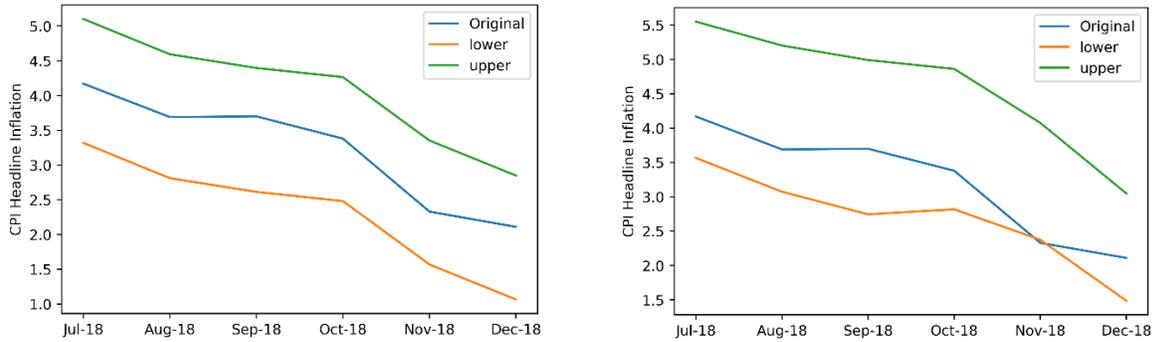

**Fig. 8.** PIs for CPI Headline using the 2-stage model (*left*) and LUBE + GD (*right*)

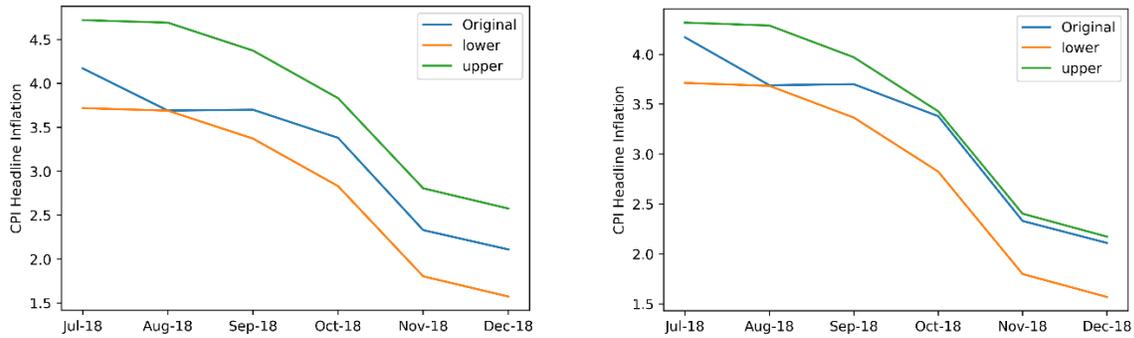

**Fig. 9.** PIs for CPI Headline using the 3-stage model (single random no.) (*left*) and the 3-stage model (two random no.) (*right*)

## *6.3. CPI Fuel and Light inflation*

Table 7 presents the PICP and PIAW values using Two-stage model (Chaos + NSGA-II{SMAPE, DS}), and three-stage model (Chaos + NSGA-II{SMAPE, DS} + NSGA-II{PICP, PIAW}). According to the results, the performance of 2-stage model is improved by 6.4% compared to LUBE + GD in w.r.t PICP metric and 35.4% in terms of PIAW metric and when compared with LUBE + LSTM, PICP showed similar performance & PIAW got improved by 34%. By running the NSGA-II second time with PICP and PIAW as objectives, the performance of the model improved by 20% in terms of PICP and got deteriorated by 6% in terms of PIAW but the performance enhancement comes with an extra computational cost of running NSGA-II second time compared to single run using 2-stage model. Fig. 10 depicts that Chaos + NSGA-II the predicted values of the two-stage model with LUBE + GD model, and it reveals that 2-stage model could yield very closer prediction intervals compared to LUBE + GD method for CPI Fuel & Light. Fig. 11 compares the performance of the two variants of the 3-stage model.

**Table 7:** Results for the CPI Fuel & Light PICP and PIAW

| Method | PICP | PIAW |
|---|---|---|
| LUBE Method with Gradient Descent | 0.78 ± 0.09 | 2.60 ± 0.49 |
| LUBE Method with LSTM [25] | 0.82 ± 0.08 | 2.58 ± 0.51 |
| NSGA-II based 2-stage model (Grid search) | 0.83 ± 0.06 | 1.68 ± 0.26 |
| NSGA-II based 3-stage model (Using single random number) | 1.00 ± 0.00 | 1.78 ± 0.33 |
| NSGA-II based 3-stage model (Using two random numbers) | 1.00 ± 0.00 | 1.82 ± 0.40 |

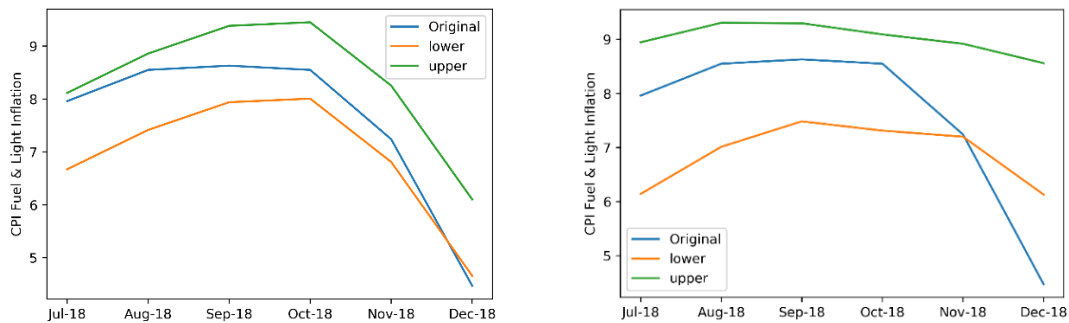

**Fig. 10.** PIs for CPI Fuel & Light using the 2-stage model (left) and LUBE + GD (right)

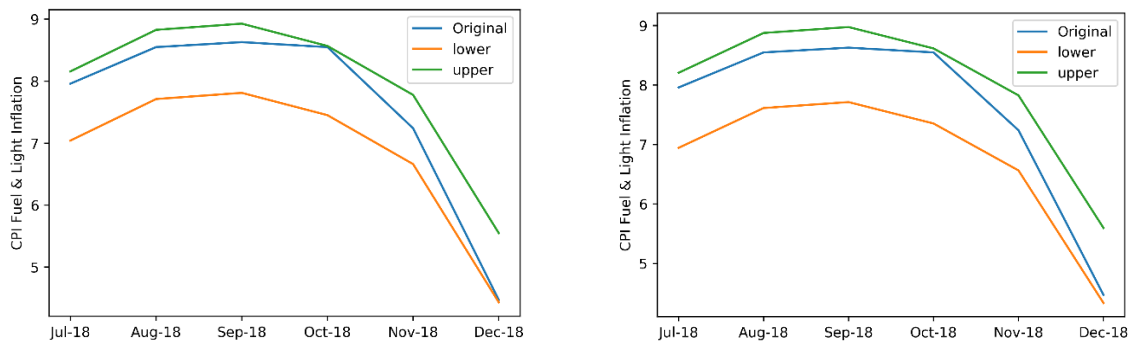

**Fig. 11.** PIs for CPI Fuel & Light using the 3-stage model (single rand no.) (*left*) and the 3-stage model (two random no.) (*right*

It is interesting to note that the NSGA-II based 2-stage and 3-stage models outperformed the LSTM based models across all the three datasets. This is primarily because the problem is now explicitly modeled as a bi-objective optimization problem and the superior exploration and exploitation skills of the NSGA-II also immensely contributed.

Now we discuss the results of the empirical attainment function (EAF), which is suggested by Fonseca et al. [49]. He argued rightly that it is a challenging task to analyze Pareto Optimal fronts for all 30 runs of an EMO, which we constructed for all datasets in our study. The EAF plot describes the probabilistic distribution of outcomes generated by the stochastic algorithm in objective space [40, 41]. The EAF plots depicted in Fig. 12 through 16 describe three types of attainment surfaces, namely the best, median and the worst of the 2-stage model and 3-stage models.

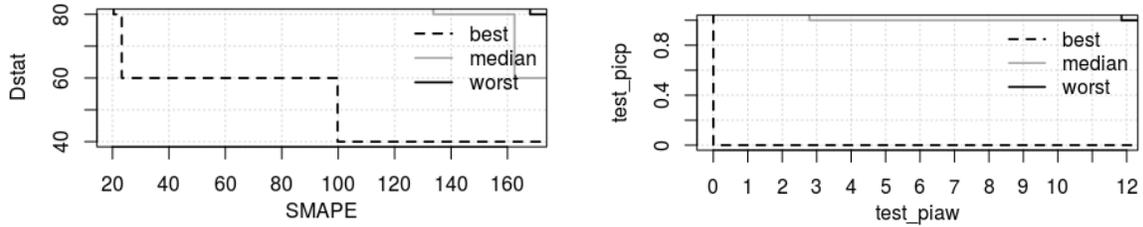

**Fig.12.** EAF plot for CPI food & beverages using the 2-stage model(left) and the 3-stage model with single random number (right).

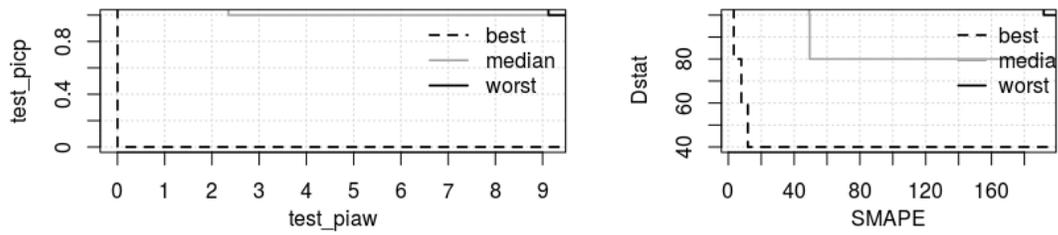

**Fig.13.** EAF plot for CPI Food & Beverages using the 3-stage model with two random numbers (Left) and CPI Headline using the 2-stage model (right)

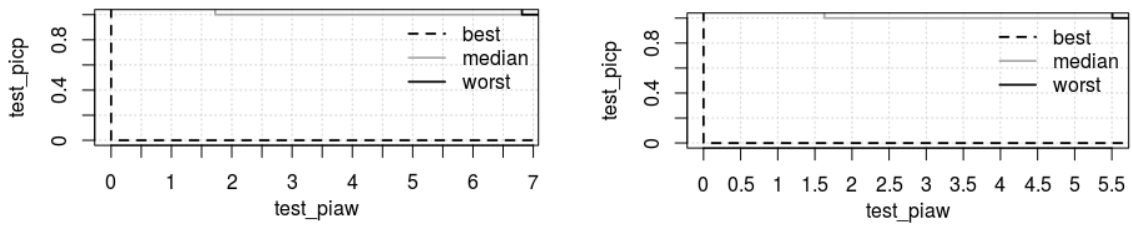

**Fig.14.** EAF plot for CPI Headline inflation using the 3-stage model with single random number (Left) and the 3-stage model with two random numbers (right)

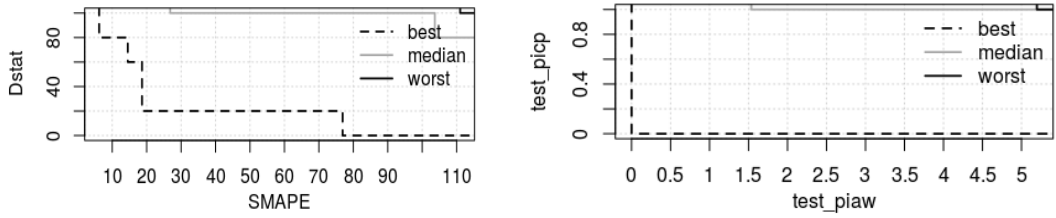

**Fig.15.** EAF for CPI Fuel & Light inflation using the 2-stage model (left) and the 3-stage model with single random number (right)

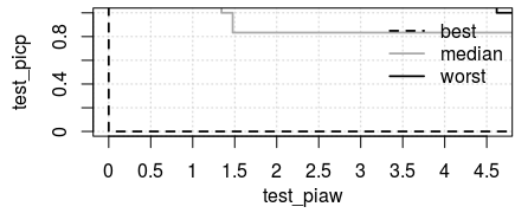

**Fig.16.** EAF for CPI Fuel & Light inflation the 3-stage model with two random numbers

**Table 8:** Hyperparameters used for different CPI data sets.

| Method | Hyper parameters | Food & Beverages | Fuel & Light | Headline |
|---|---|---|---|---|
| LUBE with GD | soften | 20 | 20 | 20 |
| | # nodes | 6 | 9 | 15 |
| | learning rate | 0.02 | 0.02 | 0.02 |
| | decay rate | 0.01 | 0.01 | 0.01 |
| NSGA-II (variant-1) | Population | 50 | 50 | 50 |
| | Crossover rate | (real_sbx, prob=0.8, eta=15) | (real_sbx, prob=0.85, eta=15) | (real_sbx, prob=0.95, eta=15) |
| | Mutation rate | (real_pm, prob=1.0, eta=20) | (real_pm, prob=1.0, eta=20) | (real_pm, prob=1.0, eta=20) |
| | # Generations | 300 | 100 | 50 |
| NSGA-II (variant-2) | Population | 90 | 90 | 90 |
| | Crossover rate | (real_sbx, prob=0.75,eta=15) | (real_sbx, prob=0.85, eta=15) | (real_sbx, prob=0.95,eta=15) |
| | Mutation rate | (real_pm, prob=1.0, eta=20) | (real_pm, prob=1.0, eta=20) | (real_pm, prob=1.0, eta=20) |
| | # Generations | 300 | 350 | 100 |
| NSGA-II (variant-3) | Population | 70 | 75 | 70 |
| | Crossover rate | (real_sbx, prob=0.75,eta=15) | (real_sbx, prob=0.8, eta=15) | (real_sbx, prob=0.75,eta=15) |
| | Mutation rate | (real_pm, prob=1.0, eta=20) | (real_pm, prob=1.0, eta=20) | (real_pm, prob=1.0, eta=20) |
| | # Generations | 200 | 300 | 400 |

## 7. Conclusions

The paper proposes a novel three-stage and two-stage models namely Chaos + NSGA-II{SMAPE, DS} & Chaos + NSGA-II{SMAPE, DS} + NSGA-II{PICP, PIAW} for generating PIs for macroeconomic time series. The results in terms of PICP and PIAW on test sets indicate that the proposed models outperformed the LUBE + GD & LUBE + LSTM. The proposed models' intuitive objective functions and low computational costs are the key advantages over LUBE method. The 3-stage models though showed improvement over the 2-stage models in terms of PICP but lag behind in terms of PIAW metric with the extra computational cost of running NSGA-II second time. Overall, the proposed model results are inspiring, and we recommend the applications of these NSGA-II based 2-stage and 3-stage models in related other economic and noneconomic time series data.